\PassOptionsToPackage{table}{xcolor}
\documentclass[10pt,twocolumn,letterpaper]{article}

\usepackage[pagenumbers]{iccv} 

\usepackage[accsupp]{axessibility}
\definecolor{iccvblue}{rgb}{0.21,0.49,0.74}
\definecolor{darkgreen}{rgb}{0, 0.5, 0}
\definecolor{myblue}{RGB}{47, 114, 193}
\definecolor{brickred}{rgb}{0.8, 0.25, 0.33}
\definecolor{brandeisblue}{rgb}{0.0, 0.44, 1.0}
\definecolor{blueish}{rgb}{0.0, 0.3, 0.6}
\usepackage[pagebackref,breaklinks,colorlinks,allcolors=iccvblue]{hyperref}
\usepackage{enumitem}

\usepackage[table]{xcolor}
\usepackage{multirow}
\usepackage{graphicx}
\usepackage{booktabs}

\usepackage{soul}
\usepackage{arydshln}
\definecolor{cGreen}{HTML}{2e75b5}
\definecolor{cgray}{HTML}{FAFAFA}
\usepackage{hyperref}
\usepackage{url}
\usepackage{caption}
\usepackage{wrapfig}
\definecolor{orange}{HTML}{cc7700}
\definecolor{green}{HTML}{339955}
\definecolor{Highlight}{rgb}{0.12,0.49,0.85}
\definecolor{my_red}{HTML}{FE4444}
\definecolor{tab_green}{RGB}{224, 254, 220}
\definecolor{tab_blue}{RGB}{217, 217, 252}
\usepackage{pifont}
\setcitestyle{circle}
\usepackage[capitalize]{cleveref}
\crefname{section}{Sec.}{Secs.}
\Crefname{section}{Section}{Sections}
\Crefname{table}{Table}{Tables}
\crefname{table}{Tab.}{Tabs.}
\Crefname{algorithm}{Algorithm}{Algorithms}
\crefname{algorithm}{Algo.}{Algos.}
\crefname{algocf}{Algo.}{Algos.}
\Crefname{algocf}{Algorithm}{Algorithms}

\usepackage{xspace}
\usepackage[ruled,vlined]{algorithm2e}
\definecolor{commentcolor}{RGB}{110,154,155}   
\newcommand{\PyComment}[1]{\ttfamily\textcolor{commentcolor}{\# #1}} 
\newcommand{\PyCode}[1]{\ttfamily\textcolor{black}{#1}} %

\newcommand{\NAME}{FastVAR\xspace}

\def\paperID{****} 
\def\confName{ICCV}
\def\confYear{2025}

\title{FastVAR: Linear Visual Autoregressive Modeling via Cached Token Pruning}

\author{
  Hang Guo$^{1}$\enspace 
  Yawei Li$^{2,*,\dagger}$\enspace 
  Taolin Zhang$^{1}$\enspace 
  Jiangshan Wang$^{1}$ \\
  Tao Dai$^{3,*}$\enspace 
  Shu-Tao Xia$^{1,4}$\enspace 
  Luca Benini$^{2}$ \\ 
  \textsuperscript{1}Tsinghua University\enspace 
  \textsuperscript{2}ETH Z\"{u}rich \enspace
  \textsuperscript{3}Shenzhen University\enspace
  \textsuperscript{4}Peng Cheng Laboratory\\
  \vspace{-4mm}
}

\begin{document}

\renewcommand{\thefootnote}{\fnsymbol{footnote}}

\twocolumn[{%
\renewcommand\twocolumn[1][]{#1}%
\maketitle
\vspace{-13mm}
\begin{center}
    \captionsetup{type=figure}
    \includegraphics[width=\linewidth]{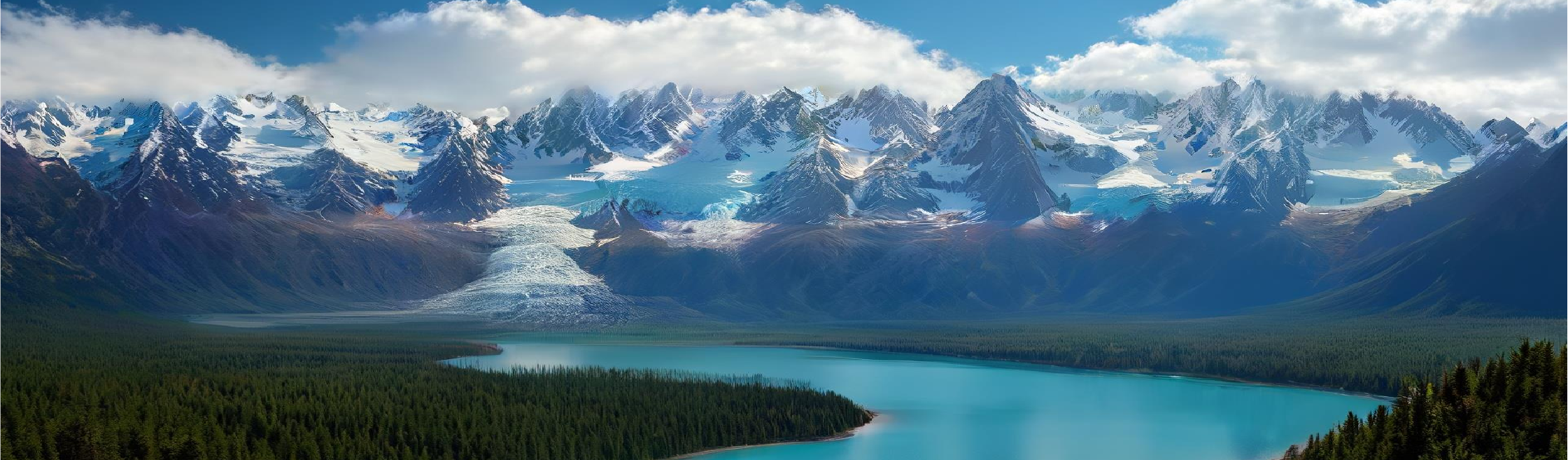}
    \vspace{-6mm}
    \captionof{figure}{Our \NAME can generate one 2K image using one NVIDIA 3090 GPU, while existing baseline fails due to out of memory.}
    \label{fig:teaser}
\end{center}%
}]

\footnotetext[1]{Corresponding Authors, $\dagger$Project Lead.}

\begin{abstract}

\vspace*{-3mm}
Visual Autoregressive (VAR) modeling has gained popularity for its shift towards next-scale prediction. However, existing VAR paradigms process the entire token map at each scale step, leading to the complexity and runtime scaling dramatically with image resolution. To address this challenge, we propose \NAME, a post-training acceleration method for efficient resolution scaling with VARs. Our key finding is that the majority of latency arises from the large-scale step where most tokens have already converged. Leveraging this observation, we develop the cached token pruning strategy that only forwards pivotal tokens for scale-specific modeling while using cached tokens from previous scale steps to restore the pruned slots. This significantly reduces the number of forwarded tokens and improves the efficiency at larger resolutions. Experiments show the proposed \NAME can further speedup FlashAttention-accelerated VAR by \textbf{2.7$\times$} with negligible performance drop of \textbf{\textless 1\%}. We further extend \NAME to zero-shot generation of higher resolution images. In particular, \NAME can generate one \textbf{2K} image with \textbf{15GB} memory footprints in \textbf{1.5s} on a single NVIDIA 3090 GPU. Code is available at \url{https://github.com/csguoh/FastVAR}.

\vspace*{-4mm}

\end{abstract}

\section{Introduction}
\label{sec:intro}

The next-token prediction of Autoregressive (AR) models~\cite{li2024mar,sun2024llamagen,xie2024showo,liu2024lumina,wang2024emu3} has demonstrated performance competitive with diffusion models for visual generation~\cite{sun2024llamagen} as well as the potential for unified vision understanding and generation~\cite{xie2024showo,wu2024vila,wang2024emu3,wu2024janus}. However, this token-by-token paradigm suffers from numerous decoding steps. Recently, Visual Autoregressive (VAR) modeling~\cite{tian2024var} has shifted the paradigm to next-scale prediction, enabling image generation in fewer steps. Under this new paradigm, some works~\cite{han2024infinity,tang2024hart} have developed VAR-based models for text-to-image generation and obtained promising results.

\begin{figure}[!t]
    \centering
    \includegraphics[width=0.99\linewidth]{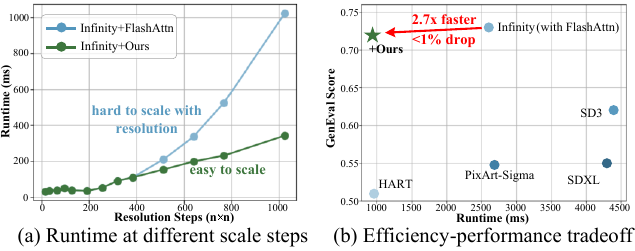}
    \vspace{-3mm}
    \caption{\NAME exhibits promising resolution scalability, and can achieve noticeable speedup with negligible performance drop.}
    \label{fig:teaser_v2}
    \vspace{-6mm}
\end{figure}

Despite their potential, existing VAR-based methods~\cite{tian2024var,han2024infinity,tang2024hart} face a critical challenge: \textit{the computational complexity and runtime latency scale dramatically with image resolution}.
Specifically, unlike next-token prediction which processes only one token per step, the next-scale prediction of VAR requires processing the entire token map at each decoding step. As a result, the number of tokens increases in $\mathcal{O}(n^2)$ with the image resolution $n\times n$, and even leads to $\mathcal{O}(n^4)$ complexity in the attention~\cite{vaswani2017attention} layer. Empirically, as shown in ~\cref{fig:teaser_v2}(a), even when FlashAttention~\cite{dao2022flashattention} is enabled, VAR models still exhibit super-linear runtime latency. 
Consequently, this significant computational complexity prevents existing VAR-based models from scaling to higher resolutions, such as 2K.

\begin{figure*}[!t]
    \centering
    \includegraphics[width=0.99\linewidth]{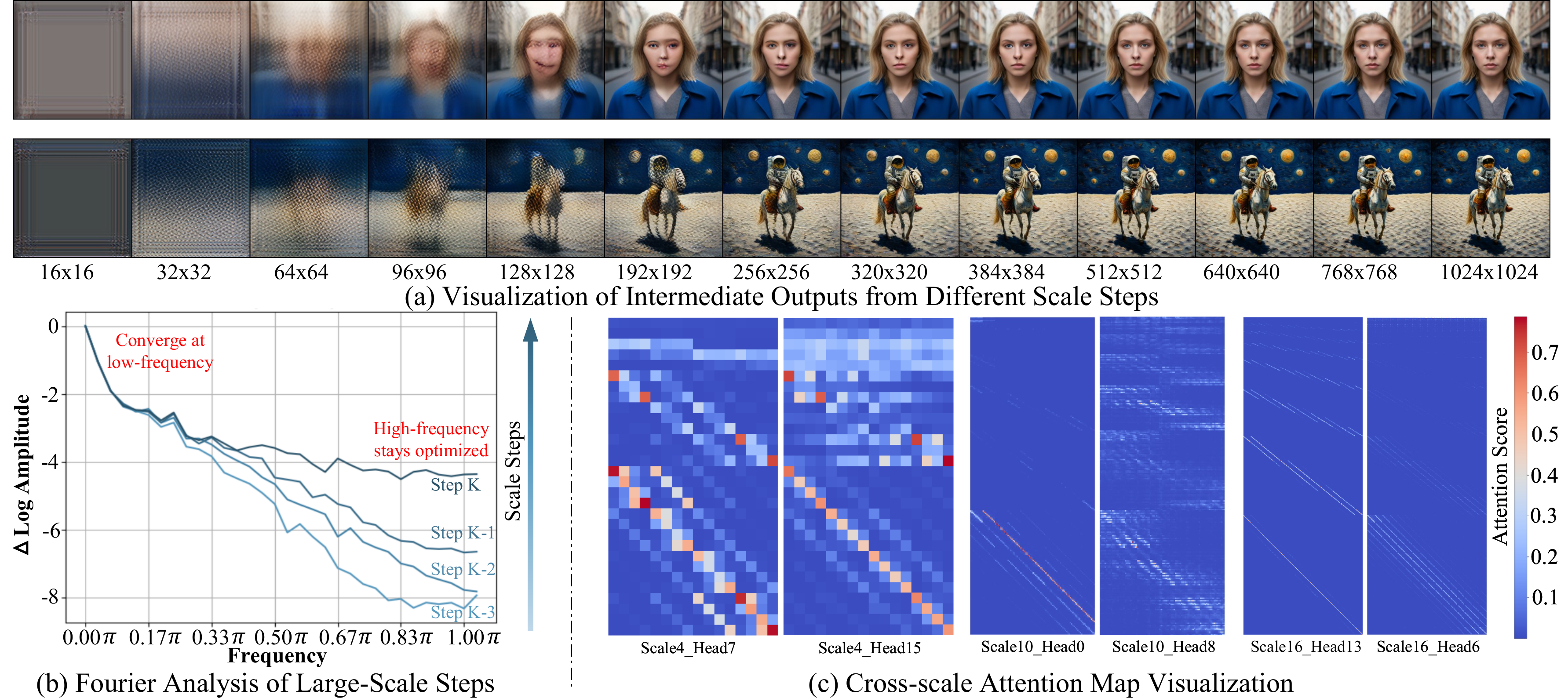}
    \vspace{-3mm}
    \caption{(a) We resize different $\Tilde{r}_k$ to the same size for better presentation. (b) Each curve represents the frequency characteristics of a certain-scale token map. (c) The row represents keys of scale steps $\{1, 2,\cdots, k\}$, and the column is queries of the $k$-th scale step.}
    \label{fig:motivation}
    \vspace{-3mm}
\end{figure*}

In this work, we aim to address the resolution scaling challenge of VARs by pruning forwarded tokens. Through an in-depth analysis of the pre-trained VAR models, we identify the following key findings. 
\textbf{\textit{1) Large-scale steps are speed bottleneck but appear robustness.}} Runtime profiling in~\cref{fig:teaser_v2}(a) finds that the last two large-scale steps account for 60\% of the total runtime. Further investigation in~\cref{tab:ablation-scale-sensitivity} reveals that VAR is more resilient to token pruning at large-scale steps than at smaller scales. Thus, we direct our efforts to the large-scale steps. 
\textbf{\textit{2) High-frequency modeling matters at large-scale steps.}} Spectrum analysis in~\cref{fig:motivation}(b) reveals that the large-scale steps are optimized to model high-frequency tokens, such as texture details, and the remaining low-frequency tokens almost converge in these steps. Consequently, we can only forward the high-frequency tokens for pruning.
\textbf{\textit{3) Tokens from different scales are related.}} 
Attention map analysis in~\cref{fig:motivation}(c) shows strong diagonal sparsity across scales, indicating that tokens attend not only to same-scale neighbors but also to those from the preceding scales. Thus, we cache tokens in previous steps to compensate for the pruned slots, preserving the 2D image lattice structure and maintaining information flow.

Based on the above observations, we propose \NAME, a post-training acceleration recipe for efficient resolution scaling with VARs. 
At its core,  \NAME employs \textbf{``cached token pruning''},  which retains only pivotal tokens at large-scale steps to reduce computational overhead while using cached token maps from early-scale steps to compensate for information loss. 
To identify the pivotal tokens, we develop Pivotal Token Selection (PTS), a frequency-based scoring mechanism. Token importance is determined by filtering out the low-frequency component estimated via the direct current component, which enables efficient frequency-based token selection directly in the spatial domain.
The selected Top-K pivotal tokens are then processed by the VAR model.
To restore pruned tokens, we propose Cached Token Restoration (CTR), which first interpolates the token map from the cached scale and then reinstates the pruned tokens by indexing the interpolated token map to the pruned location. By integrating these strategies, we present \NAME which enjoys the following benefits:
\begin{itemize}[nosep]
    \item The proposed method is training-free and plug-and-play for various VAR-based backbones. 
    \item As shown in~\cref{fig:teaser_v2}(b), \NAME can be integrated with FlashAttention, with further 2.7$\times$ speedup and \textless 1\% performance drop.
    \item In~\cref{fig:teaser}, \NAME facilitates zero-shot scaling to larger-resolution and can produce one 2K image using only 15GB memory in 1.5s on a single NVIDIA 3090 GPU. 
\end{itemize}

\section{Related Work}

\noindent
\textbf{Autoregressive Visual Generation.}
The previous autoregressive (AR) methods~\cite{zheng2022movq,razavi2019generating,lee2022autoregressive,yu2021vector} mostly adopt the next-token prediction paradigm, which treats each pixel as one token and generates pixels in a GPT or BERT style~\cite{chang2023muse,achiam2023gpt,liu2019roberta,devlin2019bert,brown2020language}. For instance, some pioneering works~\cite{van2016conditional,chen2020generative} generate pixels in raster scan order with transformer models. Later, VQVAE~\cite{van2017vqvae} and VQGAN~\cite{esser2021taming} propose to quantize image patches into discrete tokens to ease the training process. Benefiting from the scaling up~\cite{kaplan2020scaling,henighan2020scaling2}, recent works have shown promising results that the autoregressive models can achieve a more competitive performance than state-of-the-art Diffusion models~\cite{rombach2022ldm,podell2023sdxl, chen2024pixartSigma, ma2024followpose,ma2024followyouremoji, chen2023pixartAlpha}. Despite this progress, existing next-token generation pipelines struggle to efficiently synthesize high-resolution images due to the numerous decoding steps~\cite{chen2024collaborative}. More recently, the Visual Autoregressive (VAR) modeling~\cite{tian2024var}, which shifts to next-scale generation fashion to allow only a few forward steps, further advances the generation quality as well as the model efficiency. Building on top of VAR, existing methods have scaled up the model for production-level text-to-image generation. For example, HART~\cite{tang2024hart} adopts the continuous diffusion module to compensate for the quantization error. Infinity~\cite{han2024infinity} employs a bitwise tokenizer for an extremely large vector vocabulary. Despite the promising ability of VAR-based methods, they struggle to scale to larger resolutions with the increasing token numbers.

\noindent
\textbf{Efficient Visual Generation.}
To speed up diffusion models~\cite{ho2020ddpm}, many efforts have been made in recent years, including distillation~\cite{salimans2022progressive,meng2023distillation}, quantization~\cite{shang2023ptq4dm,li2023qdiffusion,guo2024intlora}, pruning~\cite{zhang2024token,bolya2023tomesd,fang2023structural,zou2024toca,wang2024attnprune}, and caching~\cite{li2023fasterdm,liu2024fastericml,ma2024deepcache,ma2024learningtocache}, which either reduce the total denoising steps or to reduce the cost of single forward. Specifically, DeepCache~\cite{ma2024deepcache} proposes to reuse the intermediate features of low-resolution layers in the U-Net. ToMeSD~\cite{bolya2023tomesd} merges similar tokens into one token with subsequent unmerging for acceleration. Unfortunately, these methods are specifically designed for diffusion, which can achieve only sub-optimal performance or cannot be used in VAR. In the efforts to accelerate AR modeling~\cite{li2024mar,sun2024llamagen,xie2024showo,liu2024lumina,wang2024emu3}, one prevalent solution is to reduce the forward step using parallel decoding strategies~\cite{he2024zipar,teng2024accelerating,santilli2023accelerating}. For instance, speculative decoding~\cite{cai2024medusa,leviathan2023fast} utilizes a small draft model to generate candidate tokens, which are then verified in parallel by the larger model. However, the well-developed decoding strategy cannot be directly applied to the next-scale paradigm, whose each step involves multiple pixels~\cite{tian2024var}. As for fast VAR models, one related work is the CoDe~\cite{chen2024collaborative}, which uses model ensemble techniques to use a small model on the costly large resolution. However, CoDe highly depends on the availability of different-sized models. In this work, we attempt to speed up the VARs by pruning extra tokens, which is training-free and is generic for multiple VAR methods.

\section{Method}

\subsection{Preliminary}

The Visual Autoregressive (VAR) modeling~\cite{tian2024var} redefines the traditional AR by shifting from a ``next-token prediction'' to a ``next-scale prediction'', where each autoregressive unit is a token map of varying scales instead of a single token. For a given image feature map, VAR first quantizes it into $K$ multi-scale token maps $\mathcal{R} = \{r_1, r_2,\cdots, r_K\}$ with progressively larger resolutions using pre-definced scale schdual \{$(h_1,w_1), (h_2,w_2), \cdots, (h_K, w_K)$\}. Then the $\mathcal{R}$ is used to train a Transformer~\cite{vaswani2017attention} to learn the joint distribution using the causal formulation:
\begin{equation}
p(r_1,r_2,\cdots,r_K)=\prod_{k=1}^{K}p(r_k | r_1,r_2,\cdots,r_{k - 1}),
\end{equation}
where the start token $r_1$ is the encoded text embedding from the LLMs~\cite{yang2024qwen2,chung2024t5} under given user prompts. 
During training, existing methods~\cite{tang2024hart,han2024infinity,tian2024var} usually employ a residual strategy to reduce learning difficulty. In detail, the model outputs at the $k$-th scale step $f_k$ is the residual of the intermediate prediction $\Tilde{r}_k$:
\begin{equation}
\Tilde{r}_k = \mathrm{interpolate}(\Tilde{r}_{k-1}, (h_k,w_k))+f_k, 
\label{eq:recursive_form}
\end{equation}
where the ``$\mathrm{interpolate}(\cdot, (h_k,w_k))$'' denotes upsample a token map to the size of $(h_k, w_k)$. Unfolding ~\cref{eq:recursive_form} derives the following cumulative form:
\begin{equation}
    \Tilde{r}_k = \sum_{i=1}^{k} \mathrm{interpolate}(f_i, (h_k,w_k)).
  \label{eq:cum_sum}
\end{equation}
In the last scale step, the predicted token maps $\Tilde{r}_K$ is used to generate the final image. During inference, since the outputs at different scales are generated autoregressively, the KV-Cache can be adopted to avoid re-computing of previous steps. 
Although the VAR paradigm can generate images within only $K$ scale steps, the Transformer needs to process all $h_k \times w_k$ tokens at the $k$-th step, hindering scaling to higher-resolution image synthesis.

\begin{figure*}[!t]
    \centering
    \includegraphics[width=0.96\linewidth]{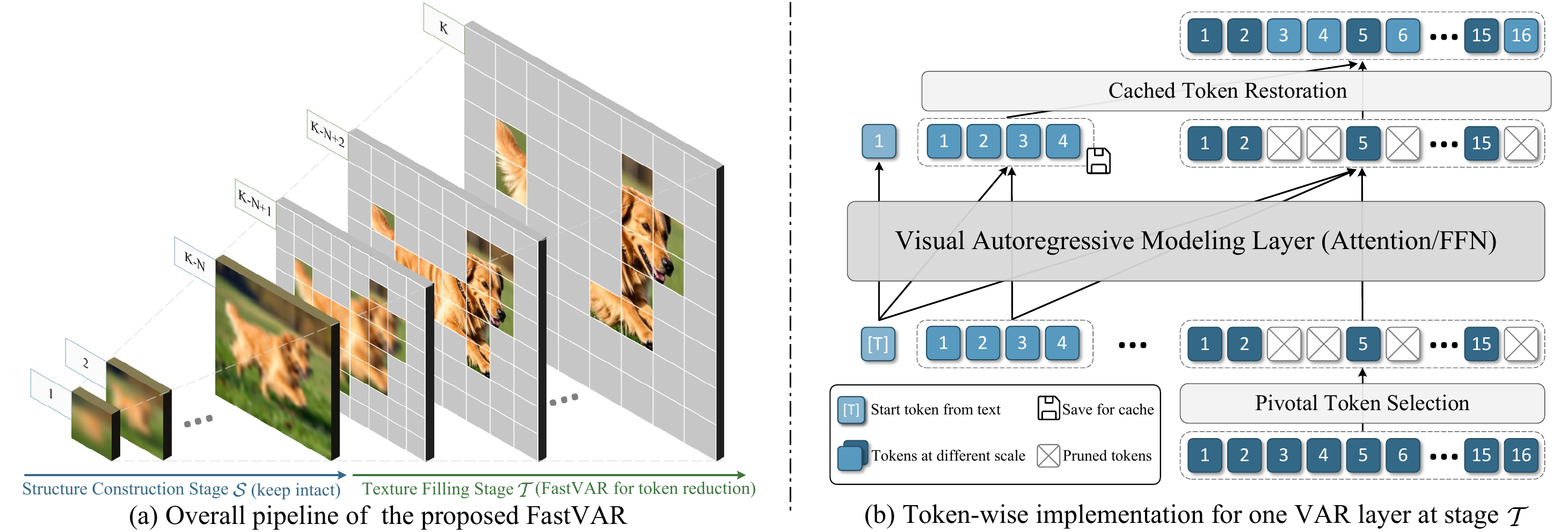}
    \vspace{-2mm}
    \caption{(a) We keep the \textbf{Structure Construction Stage} as standard VAR and additionally store token maps at the $(K-N)$-th step as the cache. Then the \textbf{Texture Filling Stage} applies \NAME to process only pivotal tokens for fast inference. (b) For Attention or FFN layers using \NAME, we first employ the Pivotal Token Selection (PTS) for token reduction. After the model forward, we develop the Cached Token Restoration (CTR) to restore the original token numbers using the caches from the previous scale step.}
    \label{fig:pipeline}
\end{figure*}

\subsection{Motivation}
\label{sec:motivation}
In this work, we aim to approach linear VAR models to advance high-resolution image generation. To this end, we investigate the scale-wise behavior of the pre-trained VAR model~\cite{han2024infinity}. We examine from three angles: the latency profiling, the specific role of each step, and the cross-scale token dependency. Conclusions are summarized as follows.

\smallskip
\noindent
\textit{\textbf{Observation 1: Large-scale steps are speed bottleneck but appear robustness.}}
As shown in the runtime curve in~\cref{fig:teaser_v2}(a), even the FlashAttention is enabled, the inference latency of the VAR exhibits a super-linear complexity with resolution steps. This is because the Transformer needs to take all the tokens at once. As a result, the tokens from the large-scale steps make up the majority of the total token length. For example, the last two steps even occupy 60\% of the total time. On the other hand, the experiments in~\cref{sec:ablation} suggest the large-scale steps are more robust to token drop than small-scale counterparts under the same pruning ratio. Therefore, it is accessible to prune some ``unimportant tokens'' at the large scale step to speed up token processing. However, challenges still remain about which tokens should be pruned and how to compensate for the information loss from these pruned tokens.

\smallskip
\noindent
\textit{\textbf{Observation 2:  High-frequency modeling matters at large-scale steps.}}
To find which token should be pruned, we visualize the intermediate predictions $\Tilde{r}_k$ in~\cref{fig:motivation}(a). Interestingly, we find that the generation process of the pre-trained VAR model can be decomposed into two stages. The Structure Construction Stage, which includes small-scale steps, is trained to generate the outline of a subject. After that, at the Texture Filling Stage, which is composed of large-scale steps, is responsible for adding details based on the previous sketch. Given that the outline is low-frequency while the texture is high-frequency, we therefore infer that the VAR mainly models the high-frequency texture in the large-scale stage while keeping the low-frequency structure almost intact. For further verification, we show in~\cref{fig:motivation}(b) that the low-frequency component almost converges in the large-scale steps, while the high-frequency modeling stays optimized with noticeable variations.
Therefore, we can prune the redundant low-frequency tokens and forward only the pivotal high-frequency tokens at the large-scale steps for token number reduction.

\smallskip
\noindent
\textit{\textbf{Observation 3: Tokens from different scales are related.}}
After reducing the input tokens for fast forward, we still need to restore the output tokens to the original numbers in order to maintain the 2D image structure. To this end, we analyze the cross-scale attention map of the unpruned VAR model, where the query at the current scale interacts with KV caches from all previous scales. As depicted in~\cref{fig:motivation}(c), we observe that the attention map not only exhibits a high diagonal score at self-scale, but also a high diagonal score at cross-scale. This behavior indicates that a token in the current scale not only attends to its same-scale locals but also exhibits strong correlation to tokens at the 
neighboring positions in previous scales. This diagonal attention sparsity thus facilitates us to use only a few tokens to estimate the original output of the pruned slots, and avoids weighting multiple tokens across the token map. Specifically, we can approximate the outputs of pruned tokens by indexing in the previous-scale token maps based on the pruning location. In this way, the information loss can be compensated thanks to this high cross-scale similarity.

\subsection{Efficient Resolution Scaling for VARs}
We instantiate the above idea and present \NAME to approximate linear VAR models. The core is the cached token pruning, which prunes tokens at large scale steps while using cache from early scales to maintain information flow.

\noindent
\textbf{Overview.}
As shown in~\cref{fig:pipeline}(a), denote the scale step set of Structure Construction Stage as $\mathcal{S}=\{1, 2,\cdots, K-N\}$, the step set of Texture Filling Stage as $\mathcal{T}=\{K-N+1, \cdots, K\}$. As stated in~\cref{sec:motivation}, the speedup of pruning token maps in $\mathcal{S}$ is insignificant while producing a noticeable performance drop, we therefore keep $\mathcal{S}$ intact as the standard VAR. In addition, we save the per-layer outputs at the last scale in $\mathcal{S}$, \textit{i.e.}, the $(K-N)$-th step, as the cache for subsequent token restoration. For the set $\mathcal{T}$, we apply \NAME to prune tokens for fast-forward, followed by the token number restoration. More details are given below.

\noindent
\textbf{Pivotal Token Selection.}
We first introduce the Pivotal Token Selection (PTS) to reduce the number of tokens. As observed in~\cref{fig:motivation}(b), the low-frequency tokens almost converge in the later scale steps. Therefore, we can feed only the high-frequency tokens and prune the remaining. However, the challenge lies in that common frequency operators, such as FFT, work in the frequency domain, making it difficult to recognize the frequency characteristics of certain tokens in the original token map. To this end, our proposed PTS adopts an approximation solution. Specifically, denote $x_k$ as the input of one layer, we estimate its low-frequency component $\bar{x}_k$ as the direct current component, which can be easily obtained via the spatially global average pooling: 
\begin{equation}
\bar{x}_{k} = \mathrm{global\_avg\_pool}(x_{k}).
\end{equation}
After that, the high-frequency component is computed as the difference between the $x_k$ and $\bar{x}_k$. The pivotal score $s_k$ is defined as the L2 norm of the high-frequency maps:
\begin{equation}
s_k = ||x_k - \bar{x}_k||_2.
\end{equation}
Subsequently, we obtain the index set $\mathcal{I}$ for pivotal token selection by keeping TopK scores in $s_k$. Finally, we perform token pruning on $x_k$ through indexing with $\mathcal{I}$ to allow the fast model forward. Notably, since the PTS reduces the number of input tokens in one transformer layer, as an additional benefit, the KV-cache is also reduced accordingly, thus optimizing the GPU footprints as well as the cross-scale attention for the subsequent scale steps.

\noindent
\textbf{Cached Token Restoration.}
The proposed PTS can effectively reduce the number of tokens. However, the visual generation tasks need to restore the original token numbers to regain the
structural 2D image. For this reason, we propose Cached Token Restoration (CTR). The main rationale stems from the diagonal attention sparsity that the token at the $k$-th scale not only attends to itself, but also exhibits strong correlation with the corresponding slot from previous $\{1,2,,\cdots,k-1\}$ steps. Therefore, our proposed CTR approximates the layer outputs in the pruned locations using the counterparts from the cached step. Formally, given the cached token map $y_{K-N}$ with size $(h_{K-N},w_{K-N})$, which is the output of the corresponding layer at step $K-N$, we first upsample it to the same size of the current scale:
\begin{equation}
y^{cache}_{k} = \mathrm{interpolate}(y_{K-N}, (h_k,w_k)).
\end{equation}
Then the value from $y^{cache}_{k}$ is scattered into the pruned slots of $y_k$ using the index set $\mathcal{I}$, to generate the restored layer output $y'_k$, which shares the same token numbers as the $x_k$. We also provide an explanation for the choice of $K-N$ as the caching step, see \textit{Suppl.} for more details.

\noindent
\textbf{Implementation of FastVAR.}
Thanks to the token number invariance of the PTS+CTR, as shown in~\cref{fig:pipeline}(b), we apply this operator pair for each Attention\&FFN layer at the large-scale steps. Furthermore, we empirically find in~\cref{sec:ablation} that the larger scale steps exhibit stronger robustness to pruning than that of early scale steps. Therefore, we develop a progressive pruning ratio schedule, in which we assign larger pruning ratios to larger scale steps in the set $\mathcal{T}$. We summarize the complete algorithm of \NAME in the \textit{Suppl.}. At last, it is worth noting that our \NAME does not access the attention map, and we show in~\cref{sec:discussion} that the proposed \NAME can be used in combination with other acceleration techniques like FlashAttention~\cite{dao2022flashattention,dao2023flashattention2} for even faster VAR generation.

\section{Experiments}

\subsection{Experimental Setup}

\noindent
\textbf{Models and Evaluations.}
We apply our \NAME on two VAR-based text-to-image models, namely HART~\cite{tang2024hart} and Infinity~\cite{han2024infinity}. Both models can generate images of up to $1024\times1024$ resolution. For a fair comparison, we keep all the hyperparameters the same as their default settings. For evaluation metrics, we compare both in terms of generation quality and inference efficiency. For generation quality, we use two popular benchmarks, \textit{i.e.}, the GenEval~\cite{ghosh2024geneval} and MJHQ30K~\cite{li2024mjhq} to validate the high-level semantic consistency and the perceptual quality, respectively. For efficiency evaluation, we adopt metrics including running time, throughputs, speedup ratio, and GPU memory costs.

\noindent
\textbf{Implementation Details.}
For the number of pruned scale steps $N$, we set $N=4$ for Infinity and $N=2$ for HART, \textit{i.e.}, only token maps from the last 4 or 2 scale steps are applied with the proposed \NAME, with other steps kept as standard VAR. For the progressive pruning ratio schedule, we set it to \{40\%, 50\%, 100\%, 100\%\} for Infinity, and \{50\%, 75\%\} for HART. The 100\% pruning ratio indicates all tokens are dropped, \textit{i.e.}, we skip the corresponding steps and interpolate the intermediate outputs to the target resolution as the final outputs. Note that this extreme ratio depends on the pruned backbones, and we provide further discussion in the \textit{Suppl.}. Unless specified, the efficiency of the unpruned baselines are already accelerated by FlashAttention~\cite{dao2022flashattention}, and we apply the proposed \NAME on top of it. 
Following the setup of existing methods~\cite{tian2024var,sun2024llamagen,chen2024collaborative}, the inference speeds for all methods are measured without including the VAE time cost since this is a shared cost for all methods. All experiments are conducted on one single NVIDIA RTX 3090 GPU with 24GB memory.

\subsection{Main Results}

\begin{table*}[!tb]
\centering 
\caption{Quantitative comparison on efficiency and quality on 1024$\times$1024 GenEval benchmarks. The marks \textcolor{blueish}{$\bullet$}, \textcolor{brickred}{$\circ$}, and \textcolor{darkgreen}{$\diamond$} denote the Diffusion, AR, and VAR-based methods, respectively. Note that the efficiency of HART and Infinity baselines are tested under FlashAttention.}
\vspace{-2mm}
\label{tab:compare-t2i}
\setlength{\tabcolsep}{6pt}
\scalebox{0.88}{
\begin{tabular}{@{}lccccccccc@{}}
\toprule
\multirow{2}{*}{Methods} & \multicolumn{5}{c}{\textbf{Inference Efficiency}} & \multicolumn{4}{c}{\textbf{Generation Quality}}          \\ \cmidrule(l){2-6}\cmidrule(l){7-10}
                         & \#Steps$\downarrow$     & Speedup$\uparrow$    & Latency$\downarrow$    & Throughput$\uparrow$   & \#Param$\downarrow$  & two\_object$\uparrow$ & position$\uparrow$ & color\_attr$\uparrow$ & Overall$\uparrow$ \\ \midrule
\textcolor{blueish}{$\bullet$}SDXL~\cite{podell2023sdxl}             & 40   & -& 4.3s  & 0.23it/s & 2.6B & 0.74 & 0.15 & 0.23 & 0.55 \\
\textcolor{blueish}{$\bullet$}PixArt-Sigma~\cite{chen2024pixartSigma}     & 20   & -& 2.7s  & 2.50it/s & 0.6B & 0.62 & 0.14 & 0.27 & 0.55 \\
\textcolor{blueish}{$\bullet$}SD3-medium~\cite{esser2024sd3}       & 28   & -& 4.4s  & 3.45it/s & 2.0B & 0.74 & 0.34 & 0.36 & 0.62 \\
\textcolor{brickred}{$\circ$}LlamaGen~\cite{sun2024llamagen}         & 1024 & -& 37.7s & 0.03it/s & 0.8B & 0.34 & 0.07 & 0.04 & 0.32 \\
\textcolor{brickred}{$\circ$}Show-o~\cite{xie2024showo}           & 1024 & -& 50.3s & 0.02it/s & 1.3B & 0.80 & 0.31 & 0.50 & 0.68 \\
\hdashline[0.5pt/2pt]
\textcolor{darkgreen}{$\diamond$}HART~\cite{tang2024hart}             & 14   &   \cellcolor{tab_green} 1.0$\times$& 0.95s & 1.05it/s & 0.7B & 0.62 & 0.13 & 0.18 & \cellcolor{tab_green} 0.51 \\
\textcolor{darkgreen}{$\diamond$}\textbf{+ FastVAR}    & 14   &      \cellcolor{tab_green}  1.5$\times$& 0.63s & 1.59it/s & 0.7B & 0.57 & 0.16 & 0.24 & \cellcolor{tab_green} 0.51 \\
\hdashline[0.5pt/2pt]
\textcolor{darkgreen}{$\diamond$}Infinity~\cite{han2024infinity}         & 13   & \cellcolor{tab_blue} 1.0$\times$  & 2.61s  & 0.38it/s & 2.0B   & 0.85 & 0.44 & 0.53 &\cellcolor{tab_blue} 0.73 \\
\textcolor{darkgreen}{$\diamond$}\textbf{+ FastVAR} & 13   &  \cellcolor{tab_blue} 2.7$\times$  & 0.95s  & 1.05it/s & 2.0B   & 0.81 & 0.45 & 0.52 & \cellcolor{tab_blue}0.72 \\ \bottomrule
\end{tabular}%
}
\end{table*}

\begin{figure*}[!t]
    \centering
    \includegraphics[width=0.98\linewidth]{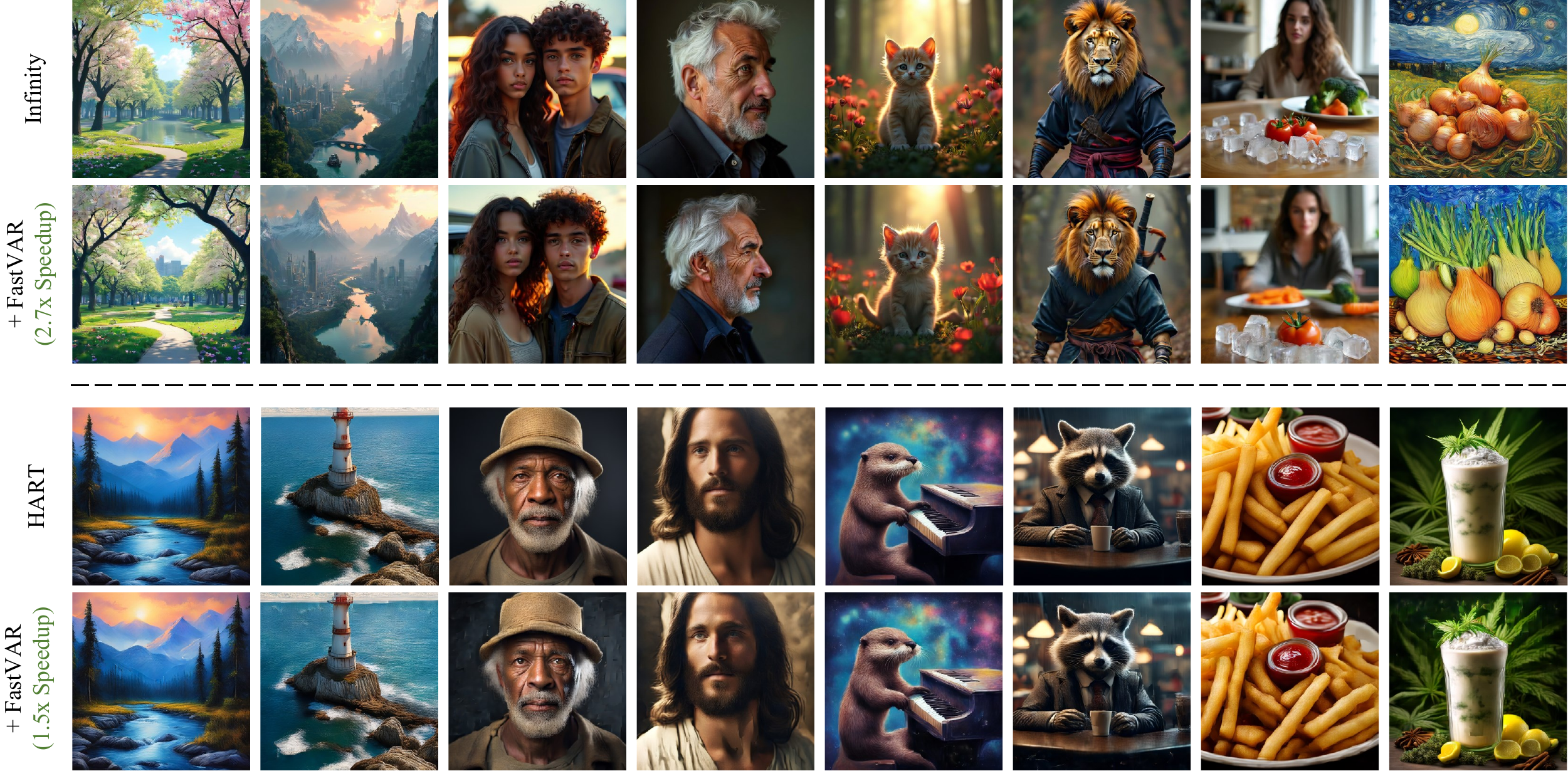}
    \vspace{-2mm}
    \caption{Qualitative comparison between the original baselines and our proposed \NAME on 1024$\times$1024 image generation. Our \NAME achieves significant speedups, \textit{e.g.}, 2.7$\times$ on Infinity~\cite{han2024infinity}, while keeping  high-quality results similar to the original model.}
    \label{fig:mjhq_compare}
    \vspace{-3mm}
\end{figure*}

\noindent
\textbf{Comparison on GenEval.}
We first evaluate the quality-efficiency trade-off on $1024\times1024$ text-to-image generation using the GenEval benchmark~\cite{ghosh2024geneval}. We compare our \NAME against various state-of-the-art methods, including Diffusion, AR, and VAR models. The results are shown in~\cref{tab:compare-t2i}. Compared with traditional AR models, our Infinity+FastVAR can achieve \textbf{even 39.7$\times$} speedup than LlamaGen~\cite{sun2024llamagen}, while achieving \textbf{125\%} performance boosts on the GenEval. Moreover, our \NAME achieves acceleration almost without performance loss, \textit{e.g.},  \textbf{1.5$\times$} speedup on the HART with the same GenEval score as the unpruned baseline. Our approach also reduces the inference GPU memory. For example, \NAME achieves \textbf{22.2\%} reduction to \textbf{14.7GB} on top of FlashAttention of 18.9GB, to produce a 1024$\times$1024 image in \textbf{0.95s}, facilitating generation on consumer-level GPUs. The above results demonstrate the generality and effectiveness of our method.

\noindent
\textbf{Comparison on MJHQ30K.}
In~\cref{tab:compare-mjhq30K}, we further validate the perceptual quality on the MJHQ30K~\cite{li2024mjhq} benchmark. It can be seen that our \NAME achieves a reasonable performance while maintaining a high speedup ratio. For instance, on the well-known challenging ``people'' category, our HART+FastVAR even achieves a FID reduction of \textbf{2.42} with \textbf{1.5$\times$} acceleration. In other categories, our method also maintains good performance with significant acceleration. We also give a qualitative visualization in~\cref{fig:mjhq_compare}, and one can see that the generated images from \NAME maintain exceptionally high quality and accurate semantic information, indicating the effectiveness of the proposed method.

\begin{figure*}[!t]
    \centering
    \includegraphics[width=0.98\linewidth]{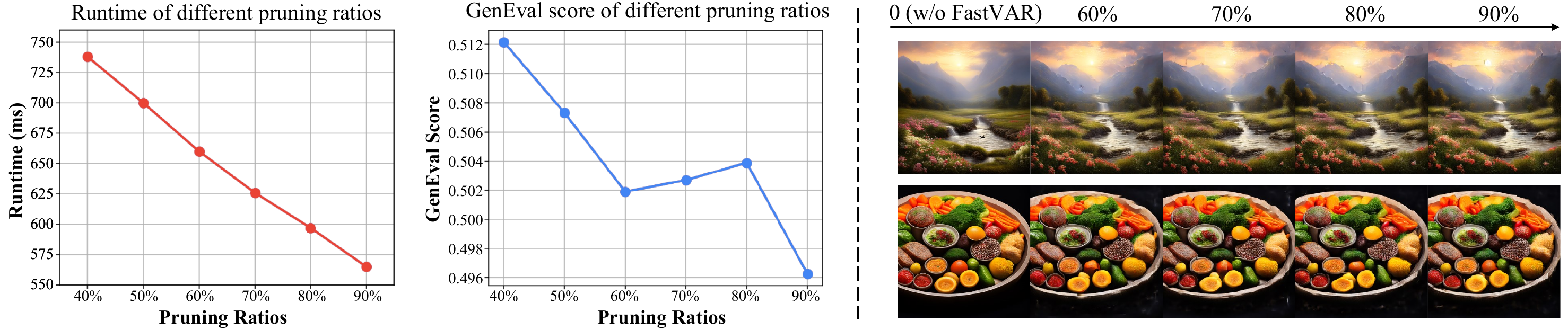}
    \vspace{-3mm}
    \caption{Quantitative and qualitative ablation experiments on different pruning ratios. We set varying pruning ratios on the last scale step of the HART model while keeping the other scale steps as the original setup.}
    \label{fig:ablation-ratio}
    \vspace{-3mm}
\end{figure*}

\begin{table}[!tb]
\centering 
\caption{Quantitative comparisons of FID and CLIP score with different generation categories on the MJHQ30K dataset.}
\label{tab:compare-mjhq30K}
\vspace{-3mm}
\setlength{\tabcolsep}{3pt}
\scalebox{0.8}{
\begin{tabular}{@{}lccccccc@{}}
\toprule
\multirow{2}{*}{Methods} & \multirow{2}{*}{Speedup} & \multicolumn{2}{c}{\textbf{landscape}} & \multicolumn{2}{c}{\textbf{people}} & \multicolumn{2}{c}{\textbf{food}} \\ \cmidrule(l){3-4}\cmidrule(l){5-6} \cmidrule(l){7-8}
                 &     & FID$\downarrow$   & CLIP$\uparrow$  & FID$\downarrow$   & CLIP$\uparrow$  & FID$\downarrow$   & CLIP$\uparrow$  \\ \midrule
SDXL~\cite{podell2023sdxl}             & -   & 30.78 & 26.35 & 35.56 & 28.01 & 35.26 & 27.98 \\
\hdashline[0.5pt/2pt]
HART~\cite{tang2024hart} & 1.0$\times$ & 25.43 & 26.82 & 30.61 & 28.47 & 30.37 & 28.03 \\
\textbf{+ FastVAR}     &  1.5$\times$   & 22.52 & 26.51 & 28.19 & 28.34 & 30.97 & 28.25 \\
\hdashline[0.5pt/2pt]
Infinity~\cite{han2024infinity}         & 1.0$\times$ & 24.68 & 26.62 & 30.27 & 27.82 & 31.55 & 26.66 \\
\textbf{+ FastVAR}  &  2.7$\times$  & 24.68 & 26.62 & 30.55 & 28.28 & 32.54 & 27.08 \\ \bottomrule
\end{tabular}%
}
\vspace{-4mm}
\end{table}

\noindent
\textbf{Difference from ToMe.}
Token Merging (ToMe)~\cite{bolya2022tome}, which merges multiple tokens into one token for efficient forward followed by token unmerging to restore the 2D shape, appears applicable to accelerate VAR backbones due to its token number invariance. In~\cref{tab:compare-tome}, we set different \NAME and ToMe variants using varying pruning ratios. One can see that ToMe fails to achieve high speedup. For example, even the 1.36$\times$ speedup can lead to noticeable FID degradation since it is difficult to compress the whole token map into limited tokens. In contrast, our \NAME can achieve 1.7$\times$ higher speedup with better FID performance.

\begin{table}[!tb]
\centering
\caption{Quantitative comparison results with ToMe~\cite{bolya2023tomesd} under different speedup settings. }
\vspace{-3mm}
\label{tab:compare-tome}
\setlength{\tabcolsep}{1pt}
\scalebox{0.78}{
\begin{tabular}{@{}lcccccc@{}}
\toprule
Methods   & Speedup$\uparrow$ & Latency$\downarrow$ & Throughput$\uparrow$ & FID$\downarrow$   & CLIP$\uparrow$  & GenEval$\uparrow$ \\ \midrule
HART~\cite{tang2024hart}      & 1.00$\times$      & 0.95s   & 1.05it/s       & 30.61 & 28.47 & 0.51    \\ \hdashline[0.5pt/2pt]
+ ToMe~\cite{bolya2023tomesd}  & 1.19$\times$    & 0.80s   & 1.25it/s       & 29.07 & 28.33 & 0.48    \\
+ ToMe~\cite{bolya2023tomesd}  & 1.36$\times$    & 0.70s   & 1.43it/s       & 35.22 & 27.99 & 0.46    \\ \hdashline[0.5pt/2pt]
+ FastVAR & 1.51$\times$    & 0.63s   & 1.59it/s       & 28.19 & 28.34 & 0.51    \\
+ FastVAR & 1.70$\times$    & 0.56s   & 1.79it/s       & 28.97 & 28.24 & 0.50    \\ \bottomrule
\end{tabular}%
}
\end{table}

\begin{table}[!tb]
\centering
\caption{Ablation experiments on the scale-wise sensitivity. We focus on the scale range [16,21,27,36,48,64] with every two consecutive scales as a pruning group. The \NAME is applied to the selected group while keeping the others unpruned.}
\label{tab:ablation-scale-sensitivity}
\vspace{-3mm}
\setlength{\tabcolsep}{6pt}
\scalebox{0.75}{
\begin{tabular}{@{}lcccccc@{}}
\toprule
scales   & ratio & Speedup$\uparrow$ & Latency$\downarrow$ & FID$\downarrow$      & CLIP$\uparrow$  & GenEval$\uparrow$ \\ \midrule
baseline & 0     & 1.00$\times$      & 0.95s   & 30.61    & 28.47 & 0.51    \\  \hdashline[0.5pt/2pt]
$[16,21]$  & 75\%  & 1.01$\times$    & 0.94s   & 34.27    & 28.43 & 0.48    \\
$[27,36]$  & 50\%   & 1.09$\times$    & 0.87s   & 27.92 & 28.49 & 0.51    \\
$[27,36]$  & 75\%  & 1.14$\times$    & 0.83s   & 29.38  & 28.56 & 0.50    \\
$[48,64]$  & 50\%  & 1.30$\times$    & 0.73s   & 27.13    & 28.43 & 0.51    \\
$[48,64]$  & 75\%  & 1.64$\times$     & 0.58s   & 28.56  & 28.36 & 0.50    \\ \bottomrule
\end{tabular}%
}
\vspace{-2mm}
\end{table}

\subsection{Ablation Studies}
\label{sec:ablation}

\noindent
\textbf{Different Pruning Ratios.} 
The pruning ratio plays an important role in balancing between efficiency and performance. Here, we conduct ablation experiments to investigate the impact of different pruning ratios. Both the quantitative and qualitative results are given in~\cref{fig:ablation-ratio}. Intuitively, increasing the pruning ratio leads to a stable runtime reduction since the model only needs to process fewer tokens. However, an over-large pruning ratio can also bring performance degradation due to the information loss of pivotal high-frequency tokens. Furthermore, the quantitative visualization suggests that increasing the pruning ratio makes certain textures and details discontinuous since the cache from the previous scale is not always an optimal approximation of the pruned tokens. Therefore, a moderate pruning ratio, \textit{e.g.}, 40\%-75\%, is adopted for HART+FastVAR to strike a sweet spot between performance and efficiency.

\noindent
\textbf{Scale-wise Sensitivity.}
As discussed in~\cref{sec:motivation}, we only focus on the token pruning of the last few scale steps. Here, we conduct ablation experiments in~\cref{tab:ablation-scale-sensitivity} to justify the rationality. It can be seen that the acceleration from pruning at an early scale steps is very limited due to the small token map size. For example, even reducing 50\% tokens at the 48 and 64 scale steps is significantly faster than that of 75\% at 16 and 21 scale steps. In addition, pruning at early scales also leads to a noticeable performance drop on both perceptual quality and semantic consistency. For instance, pruning at 48 and 64 scale steps brings 5.71 lower FID compared to that of 16 and 21 scale steps. This is because token reduction at small-scale steps disrupts the structure construction stage, and the inaccurate subject structure can exacerbate errors in subsequent scale steps.

\begin{figure}[!t]
    \centering
    \includegraphics[width=0.98\linewidth]{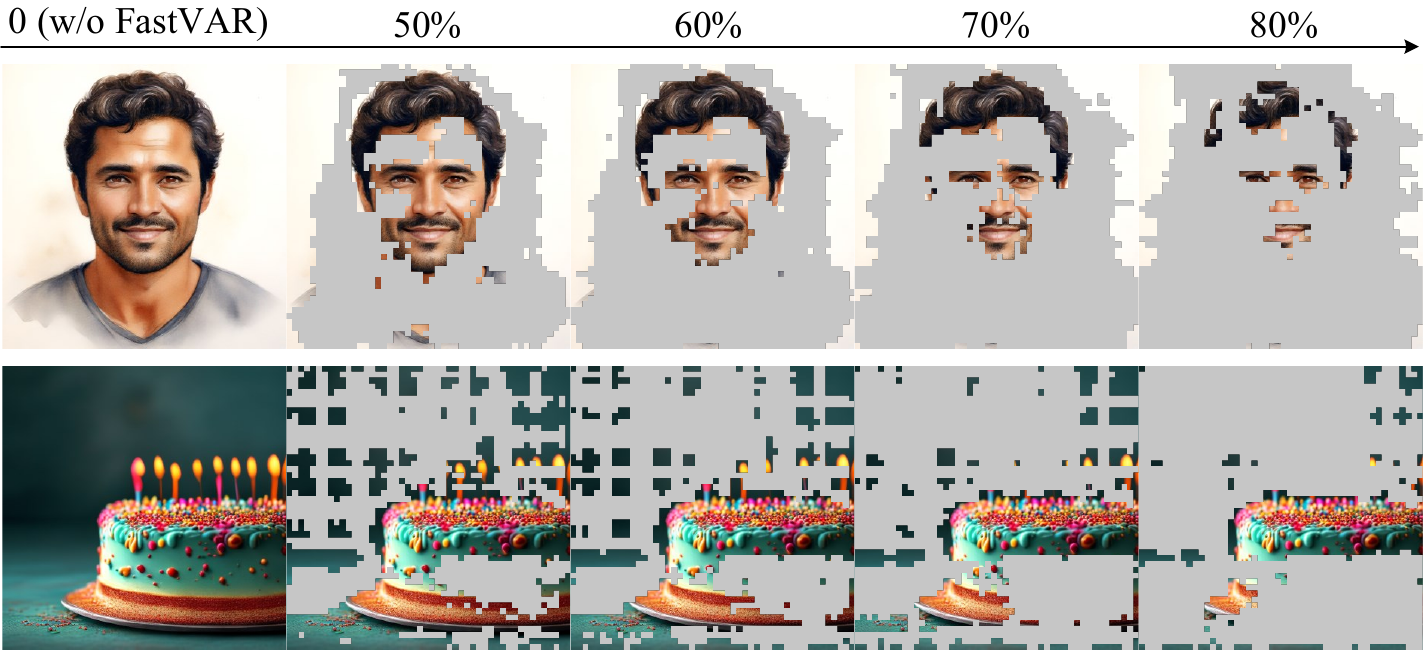}
    \vspace{-2mm}
    \caption{Visualization of pruned tokens with different ratios.}
    \label{fig:ablation-mask}
    \vspace{-3mm}
\end{figure}

\begin{figure*}[!t]
    \centering
    \includegraphics[width=0.99\linewidth]{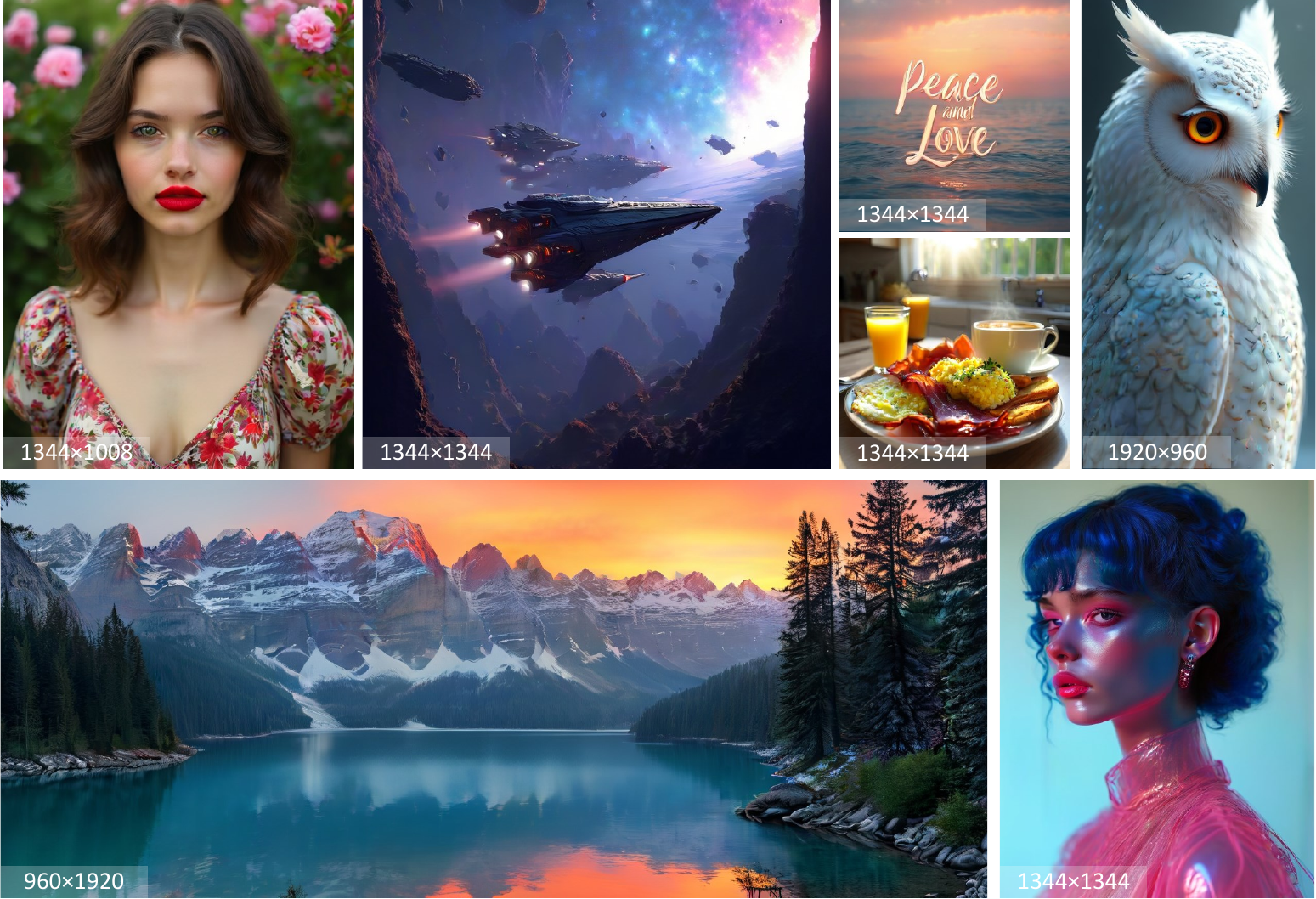}
    \vspace{-2mm}
    \caption{We extend the scale schedule of pre-trained VAR model~\cite{han2024infinity} for zero-shot larger resolution generation. We chose a modest extended scale since a larger one is observed quality drop due to the resolution gap. The images are re-scaled for better presentation.}
    \label{fig:higher-resolution}
    \vspace{-3mm}
\end{figure*}

\noindent
\textbf{Visualization of Pruned Slots.}
In~\cref{fig:ablation-mask}, we visualize the selected TopK pivotal tokens under different pruning ratios. For a given pruning ratio, \textit{e.g.}, 50\%, our \NAME prioritizes the retention of high-frequency edges or texture slots, and removes tokens from flat regions that have already converged at early scale steps. Furthermore, the importance order within the selected tokens is also meaningful. For example, in the first row, when increasing the pruning ratio from 50\% to 90\%, the human eyes, hair and mouth regions, which are the most detailed, are consistently preserved while the low-frequency cheek regions are gradually pruned. Given that our \NAME is gradient-free, this quantitative result confirms that the proposed frequency-based pivotal token selection is a reasonable heuristic.

\subsection{Discussion}
\label{sec:discussion}

\noindent
\textbf{In Conjunction with FlashAttention.}
Existing token pruning methods~\cite{zhang2023h2o,ge2023fastgen,fu2024lazyllm,xiao2023attnsink,zou2024toca} often depend on the attention map for token importance, leading to difficulties in combination with FlashAttention~\cite{dao2022flashattention} in which the attention map is inaccessible. In comparison, our \NAME can be seamlessly integrated with FlashAttention to facilitate more efficient inference. In~\cref{tab:effciiency_flashattn}, we show that using the FastVAR-only can obtain even \textbf{2.1$\times$} speedup against FlashAttention only, with \textbf{20.4\%} GPU memory reduction. After combining both, the resulting version leads to \textbf{2.7$\times$} speedup than FlashAttention only. This experiment validates that our \NAME can be orthogonally combined with other methods for further acceleration.

\noindent
\textbf{Towards Larger Resolution.}
As presented in~\cref{sec:motivation}, the original VAR struggles to scale to higher-resolution due to the increasing token numbers. Our \NAME can facilitate efficient resolution scaling, offering the potential to generate images at a larger resolution. To this end, we apply our \NAME to the Infinity~\cite{han2024infinity} for zero-shot larger resolution synthesis by appending additional steps. The results are shown in~\cref{fig:higher-resolution}. It is noteworthy that even the FlashAttention accelerated baseline is \textbf{out of memory} on a 24GB NVIDIA 3090 GPU. In contrast, our FastVAR costs only \textbf{15GB} memory and \textbf{1.3s} runtime to generate a 1344$\times$1344 image. Moreover, the generated images are of high quality, suggesting that our \NAME can facilitate production-level image generation on consumer-level GPUs.

\begin{table}[!tb]
\centering
\caption{Efficiency comparison with FlashAttn on 1024$\times$1024 image generation.}
\label{tab:effciiency_flashattn}
\vspace{-3mm}
\setlength{\tabcolsep}{3.5pt}
\scalebox{0.8}{
\begin{tabular}{lcccc}
\toprule
setups           & Speedup$\uparrow$ & Latency$\downarrow$ & Throughput$\uparrow$ & Memory$\downarrow$ \\ \midrule
SlowAttn~\cite{vaswani2017attention}   & -  & -   & -   & OOM    \\
FlashAttn only~\cite{dao2022flashattention} & 1.0$\times$ & 2.61s   & 0.38it/s    & 16.1GB \\
SlowAttn+\NAME       & 2.1$\times$  & 1.25s   & 0.80it/s   & 12.8GB \\
 \hdashline[0.5pt/2pt]
FlashAttn+\NAME & 2.7$\times$  & 0.95s   & 1.05it/s    & 11.9GB \\ \bottomrule
\end{tabular}%
}
\vspace{-3mm}
\end{table}

\section{Conclusion}
This work presents \NAME to advance high-resolution image synthesis with VARs by addressing the resolution scaling challenge. We reveal that the main latency bottleneck is the large-scale steps at which the low-frequency tokens have almost converged. To this end, we develop the cached token pruning technique to allow only process pivotal high-frequency tokens and use caches from the previous scale steps to restore the original token numbers. Thanks to the reduced forwarded tokens, our \NAME approximates linear complexity, further enabling generation at a larger resolution such as 2K. Extensive experiments validate our \NAME as an effective and generic solution for efficient resolution scaling of VAR models.

{
    \small
    \bibliographystyle{ieeenat_fullname}
    \bibliography{main}
}

\clearpage
\maketitlesupplementary
\appendix
\renewcommand{\theequation}{A.\arabic{equation}}
\setcounter{equation}{0}
\renewcommand{\thetable}{A.\arabic{table}}
\setcounter{table}{0}
\renewcommand{\thefigure}{A.\arabic{figure}}
\setcounter{figure}{0}

\section{Results on ImageNet with VAR}
In the main paper, we focus on the performance of our FastVAR on high-resolution image generation tasks. As stated in Section 3.2, applying token pruning 
on early small scale steps can lead to performance degradation due to the interference of the structure construction. Given that pruning on small token maps can not bring significant speedups, we thus do not design over-complex algorithms to further accelerate small-scale steps. However, for the sake of experimental completeness, we give the results of our \NAME on the 256$\times$256 class conditional generation on ImageNet in~\cref{tab:suppl_compare_code_var}. It can be seen that our \NAME can achieve very competitive performance against existing methods, while maintaining a high speedup ratio. Since the VAE in the existing VAR methods adopts a high compression rate, \textit{e.g.}, 16$\times$ downsample, the token map resolution at the largest scale in the VAR model~\cite{tian2024var} is only 16$\times$16 for the 256$\times$256 image generation. As a result, token pruning on this small-scale generation is much less robust compared to the larger resolution, \textit{e.g.}, 1024$\times$1024 resolution. We leave it for future work to design more generalized strategies to further include token maps at small scales.

\section{Further Efficiency Profiling}
\begin{figure}[!t]
    \centering
    \includegraphics[width=0.88\linewidth]{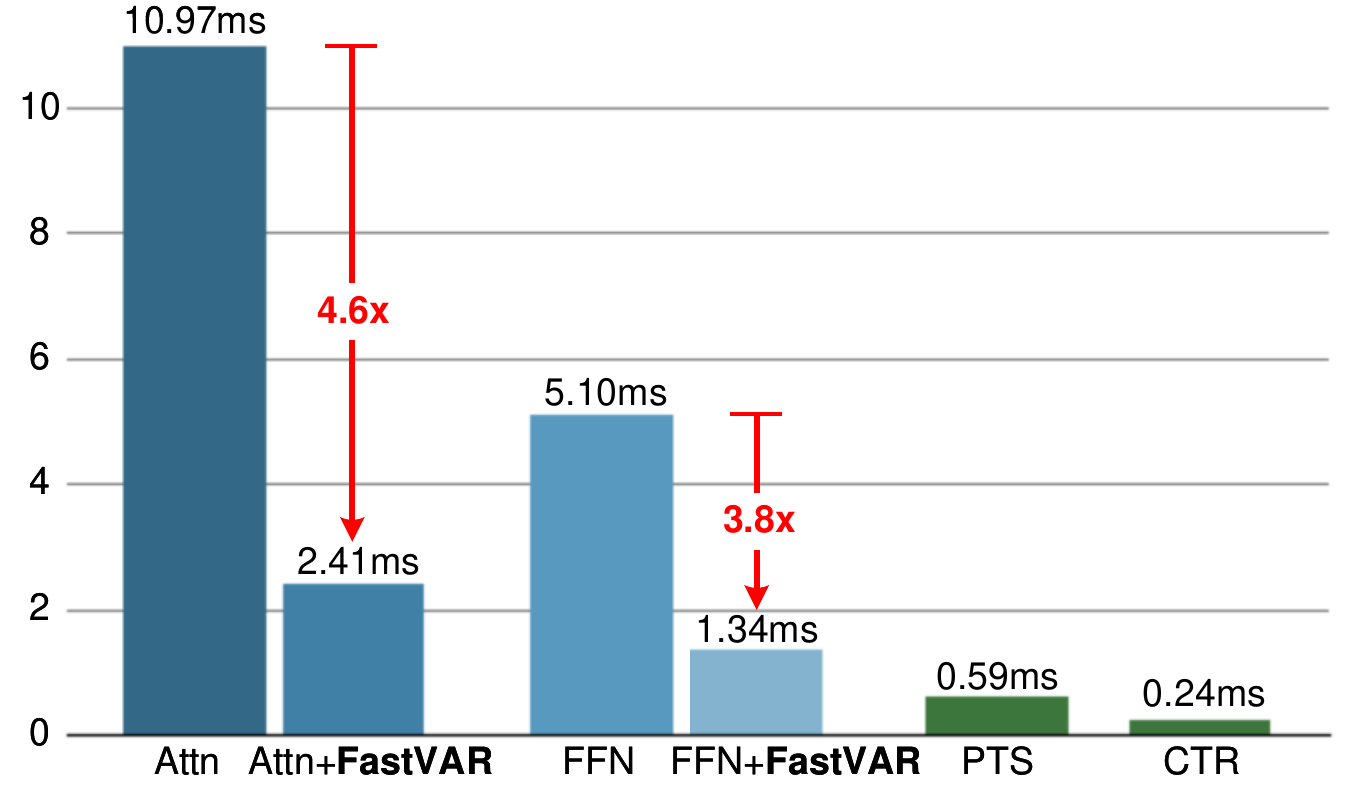}
    \vspace{-2mm}
    \caption{Efficiency Profiling of different modules in one Transformer layer. Note that the runtime of the Attn and FFN baseline is evaluated using FlashAttention.}
    \label{fig:suppl-efficiency}

\end{figure}

As demonstrated in the experiments, our \NAME can achieve significant speedup without performance degradation, \textit{e.g.}, 1.5$\times$ speedup for the HART backbone. However, this speedup ratio still shares some similar latencies as the unpruned benchmark, such as the forward pass at small scale steps. Here, we give a more fine-grained speedup comparison by directly comparing the attention and FFN under the condition of with and without the proposed \NAME. As illustrated in~\cref{fig:suppl-efficiency}, \NAME (ratio=75\%) can bring even a 4.6$\times$ speedup for the attention and a 3.8$\times$ speedup for the FFN. This result demonstrates a promising speedup upper bound of our \NAME.

Furthermore, compared to the runtime of the standard benchmark, our \NAME adds additional token importance calculation, as well as token number restoration, which may introduce additional time. Here, we give experimental results to validate the efficiency of our \NAME. As shown in~\cref{fig:suppl-efficiency}, the additional computational cost accompanying our \NAME is almost negligible. For example, the proposed PTS occupies only 0.59 ms, while the CTR occupies 0.24 ms. Thus the total additional latency from our \NAME, \textit{i.e.}, 0.63 ms, occupies only 5\% of the original attention module, which is significantly lower than the speedup brought from \NAME.

\begin{table}[!tb]
\centering
\caption{Quantitative comparison on 256$\times$256 generation on ImageNet.}
\vspace{-3mm}
\label{tab:suppl_compare_code_var}
\setlength{\tabcolsep}{1pt}
\scalebox{0.8}{
\begin{tabular}{@{}lccccccc@{}}
\toprule
Methods    & \#param & runtime  & memory & IS$\uparrow$     & FID$\downarrow$  & Precision$\uparrow$ & Recall$\uparrow$ \\ \midrule
VAR(d=24)~\cite{tian2024var}  & 1.0B    & 1.2s    & 14GB   & 313.7 & 2.29 & 82.50     & 57.45  \\
VAR(d=30)~\cite{tian2024var}  & 2.0B    & 2.2s   & 22GB   & 306.6 & 2.05 & 81.76     & 58.20  \\
CoDe(N=8)~\cite{chen2024collaborative}   & 2.3B    & 1.7s  & 15GB   & 300.4 & 2.26 & 81.31     & 58.63  \\
CoDe(N=9)~\cite{chen2024collaborative}   & 2.3B    & 2.2s    & 16GB   & 297.2 & 2.16 & 81.07     & 59.03  \\
\hdashline[0.5pt/2pt]
FastVAR(d=24) & 1.0B    & 1.1s   & 13GB   & 287.4 & 2.64 & 79.76     & 58.22  \\
FastVAR(d=30) & 2.0B    & 1.9s   & 15GB   & 288.7 & 2.30 & 80.72     & 58.64  \\ \bottomrule
\end{tabular}%
}
\end{table}

\section{Comparison on More Benchmarks}
In the main paper, we compare our FastVAR with different methods on the Geneval~\cite{ghosh2024geneval} and MJHQ30K~\cite{li2024mjhq} datasets. In order to provide a systematic evaluation, we further compare different methods on more benchmarks including HPSv2.1~\cite{wu2023hpsv21}, and ImageReward~\cite{xu2023imagereward}. The experimental results are given in~\cref{tab:suppl-more-benchmark}. It can be seen that our FastVAR maintains consistently favorable performance than other competitive token pruning baseline, while allowing for significant speedup than the original backbones. For instance, FastVAR achieves 0.81 higher HPSv2.1 score than ToMe~\cite{bolya2022tome} while being more efficient. The above results on more evaluation benchmarks further demonstrate the effectiveness of our FastVAR.

\begin{table}[!t]
\centering
\caption{More evaluation results on HPSv2.1~\cite{wu2023hpsv21} and ImageReward~\cite{xu2023imagereward} benchmarks.}
\label{tab:suppl-more-benchmark}
\vspace{-3mm}
\setlength{\tabcolsep}{3.5pt}
\scalebox{0.85}{
\begin{tabular}{l|ccc|ccc}
\hline
benchmark   & HART   & ToMe   & \multicolumn{1}{l|}{Ours}   & Infinity & ToMe   & Ours \\ \hline
Latency$\downarrow$     & 950ms & 800ms   & 630ms                        &  2600ms    &  1130ms  & 950ms     \\
Speedup$\uparrow$    & 1.0$\times$ & 1.2$\times$  & 1.5$\times$   & 1.0$\times$    &  2.3$\times$ &  2.7$\times$   \\
HPSv2.1$\uparrow$     & 28.75  & 27.04  & 27.85                       & 30.36    & 29.85  & 30.03         \\
ImageReward$\uparrow$ & 0.5658 & 0.4988 & \multicolumn{1}{l|}{0.5370} & 0.9245   & 0.8992 & 0.9129        \\ \hline
\end{tabular}%
}
\end{table}

\section{Ablation on Caching Step}
In the proposed Cached Token Restoration (CTR), we use the token map from the last scale step of the Structure Construction Stage $\mathcal{S}$, \textit{i.e.}, the $(N-K)$-th step, as the cache, which will be used to restore the original token numbers during token pruning.  Here, we conduct ablation experiments to justify the rationality by setting different scale steps as the caching step. The results are shown in~\cref{tab:suppl_cache_step}. It can be seen that setting the last element in $\mathcal{S}$ as the caching step achieves consistently the best results on all evaluation metrics. Notably, this setup has almost no performance degradation compared to the unpruned baseline models. In addition, we observe a steady performance drop when the caching step gradually moves small. This is because we use the cached token map to approximate the pruned tokens, so the gap between the last element in $\mathcal{S}$ and the steps in $\mathcal{T}$ is the smallest. Therefore, using the step that is closer to the pruned scale steps as the caching step can achieve better performance.

\section{Discussion on Extreme Pruning Ratios}
In the main paper, we mentioned that different backbones exhibit different levels of tolerance for the pruning ratio. For example, we used an even 100\% ratio for the last two scale steps of the Infinite model~\cite{han2024infinity}. However, we point out that this extreme pruning ratio does not apply to HART model~\cite{tang2024hart}. Specifically, we apply the $N=2$ and \{50\%, 100\%\} \NAME to the HART model. The experimental results are shown in~\cref{tab:suppl_extreme_ratios}. It can be seen that the extreme pruning ratio produces serious performance degradation for HART. We argue that this is due to the difference in the pre-trained model size. Specifically, the size of Infinite 2B is significantly larger than the 700M of HART. The stronger capabilities of the larger model allow for modeling more challenging textures in the earlier scale steps. As a contrast, the smaller model relies on test-time scaling~\cite{snell2025scaling} to use longer scale steps to produce complex details, and thus suffers from severe degradation when extreme pruning is applied on the last few steps.

\begin{table}[!tb]
\centering
\caption{Ablation experiments of applying extreme pruning ratios to other VAR backbone HART~\cite{tang2024hart}. We set $N=2$ in all setups, \textit{i.e.}, only the last two scale steps are pruned with \NAME.}
\vspace{-2mm}
\label{tab:suppl_extreme_ratios}
\setlength{\tabcolsep}{1.5pt}
\scalebox{0.8}{
\begin{tabular}{@{}lcccccc@{}}
\toprule
ratios       & Speedup$\uparrow$   & Latency$\downarrow$ & Throughput$\uparrow$ & FID$\downarrow$    & CLIP$\uparrow$  & GenEval$\uparrow$ \\ \midrule
no\_pruning  & 1.0$\times$      & 0.95s   & 1.05       & 30.61  & 28.47 & 0.51    \\
\{50\%,75\%\}  & 1.5$\times$ & 0.63s   & 1.59       & 28.19  & 28.34 & 0.51    \\
\{50\%,100\%\} & 1.9$\times$ & 0.51s   & 1.96       & 48.54 & 28.46 & 0.48    \\ \bottomrule
\end{tabular}%
}
\end{table}

\begin{table}[!tb]
\centering
\caption{Ablation experiments of different caching scale steps.}
\label{tab:suppl_cache_step}
\vspace{-2mm}
\setlength{\tabcolsep}{4pt}
\scalebox{0.73}{
\begin{tabular}{@{}lcccccc@{}}
\toprule
\multirow{2}{*}{cached steps} & \multicolumn{4}{c}{\textbf{GenEval}}                    & \multicolumn{2}{c}{\textbf{MJHQ30K}} \\ \cmidrule(l){2-5}\cmidrule(l){6-7}
                             & two\_object$\uparrow$ & position$\uparrow$ & color\_attr$\uparrow$ & Overall$\uparrow$ & FID$\downarrow$          & CLIP$\uparrow$         \\ \midrule
no\_pruning  &0.62& 0.13 &0.18 &0.51  &    30.61& 28.47     \\
\hdashline[0.5pt/2pt]
K-N-3                       & 0.53        & 0.11     & 0.15        & 0.47    & 40.61        & 27.36        \\
K-N-2                       & 0.60        & 0.13     & 0.19        & 0.50    & 32.39        & 27.94        \\
K-N-1                       & 0.57        & 0.13     & 0.20        & 0.49    & 29.83        & 28.25        \\ \hdashline[0.5pt/2pt]
K-N                      & 0.57        & 0.16     & 0.24        & 0.51    & 28.19        & 28.34        \\ \bottomrule
\end{tabular}%
}
\end{table}

\section{Limitation and Future Work}

Our \NAME can effectively alleviate the quadratically increasing complexity with scales, benefiting from the proposed cached token pruning. Nonetheless, 
our work can be further improved in the future in the following aspects. First, the proposed \NAME focuses mainly on the acceleration of the large-scale step which occupies the main inference time. Therefore, our method can be further improved in accelerating small-resolution image generation tasks by designing more generalized pruning strategies to include pruning small-scale token maps as well. Second, we have revealed the scale-wise sensitivity of pre-trained VAR models, \textit{i.e.}, large-scale steps are more robust to small-scale steps for pruning, which inspires us to adopt a progressive pruning ratio schedule. Therefore, utilizing more fine-grained pruning prior, \textit{e.g.}, layer-wise or even developing adaptive pruning ratios, is promising to achieve higher speedup ratios. Third, we show that the current \NAME can be combined with Flash Attention to achieve combined speedup. As other potential work on accelerating VAR models emerges, such as network quantization or fewer decoding steps, our \NAME could potentially integrate with these approaches to achieve further acceleration.

\section{Algorithm of FastVAR}

In~\cref{alg:code_box}, we give the Pytorch-like pseudocode of the proposed \NAME. Thanks to the simplicity and generality, our proposed \NAME can be seamlessly integrated into various VAR models using a few code lines.

\begin{algorithm*}[!tb]
    \caption{The pseudo-code of FastVAR algorithm, Pytorch-like}
    \label{alg:code_box}
    \small
    \SetAlgoLined
    \PyCode{def pivotal\_token\_selection(x, topk):} \\
    \Indp %
        \PyComment{calculate the direct-through component} \\ 
        pool\_x = rearrange(x, 'b (h w) c -> b c h w')\\
        pool\_x = adaptive\_avg\_pool2d(x, (1, 1))  \\
        pool\_x = rearrange(pool\_x, 'b c 1 1 -> b 1 c')  \\
        score = sum((x - pool\_x)**2, dim=-1) \\
        \PyComment{select the topK high frequency tokens}\\
        pivotal\_idx = argsort(score, dim=1, descending=True)[:, :topk, :] \\
        return gather(x, dim=1, index=pivotal\_idx) \\        
    \Indm %
    def cached\_token\_restoration(x, cache): \\
     \Indp %
        \PyComment{up-sample cache features to the size of x}\\
        restored\_x = interpolate(cache) \\
        restored\_x = rearrange(restored\_x, 'b c h w -> b h w c')\\
        \PyComment{fuse the cached and the current tokens}\\
        restored\_x.scatter\_(dim=1, index=pivotal\_idx, src=x)\\
        return restored\_x \\
    \Indm %
\end{algorithm*}

\section{More Visual Results}

\begin{figure*}[!tb]
    \centering
    \includegraphics[width=0.98\linewidth]{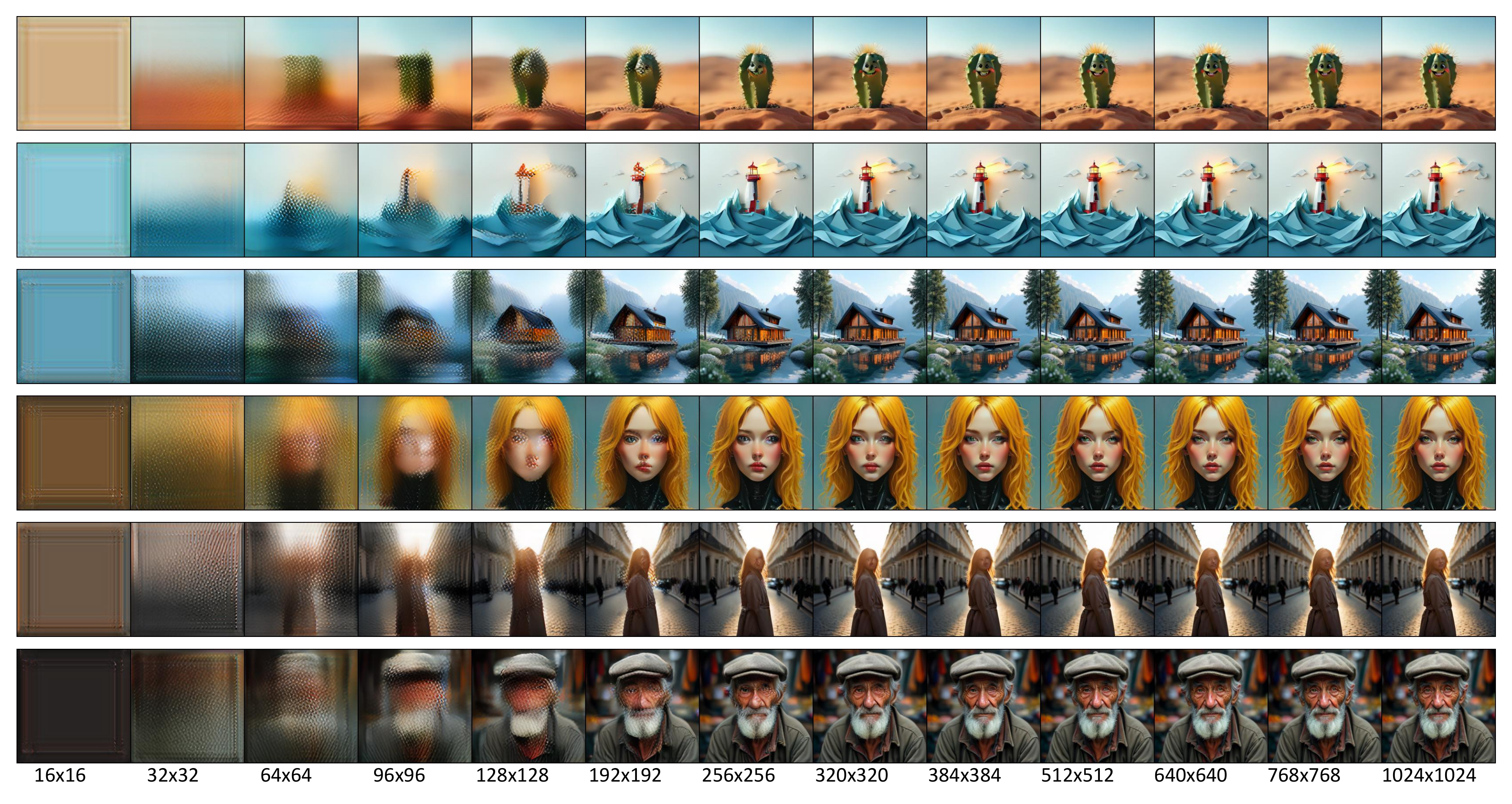}
    \vspace{-2mm}
    \caption{More visualization results of the intermediate outputs at different scale steps of pre-trained VAR model~\cite{han2024infinity}.}
    \label{fig:suppl-intermidiate}
\end{figure*}

\begin{figure*}[ht]
    \centering
    \includegraphics[width=0.96\linewidth]{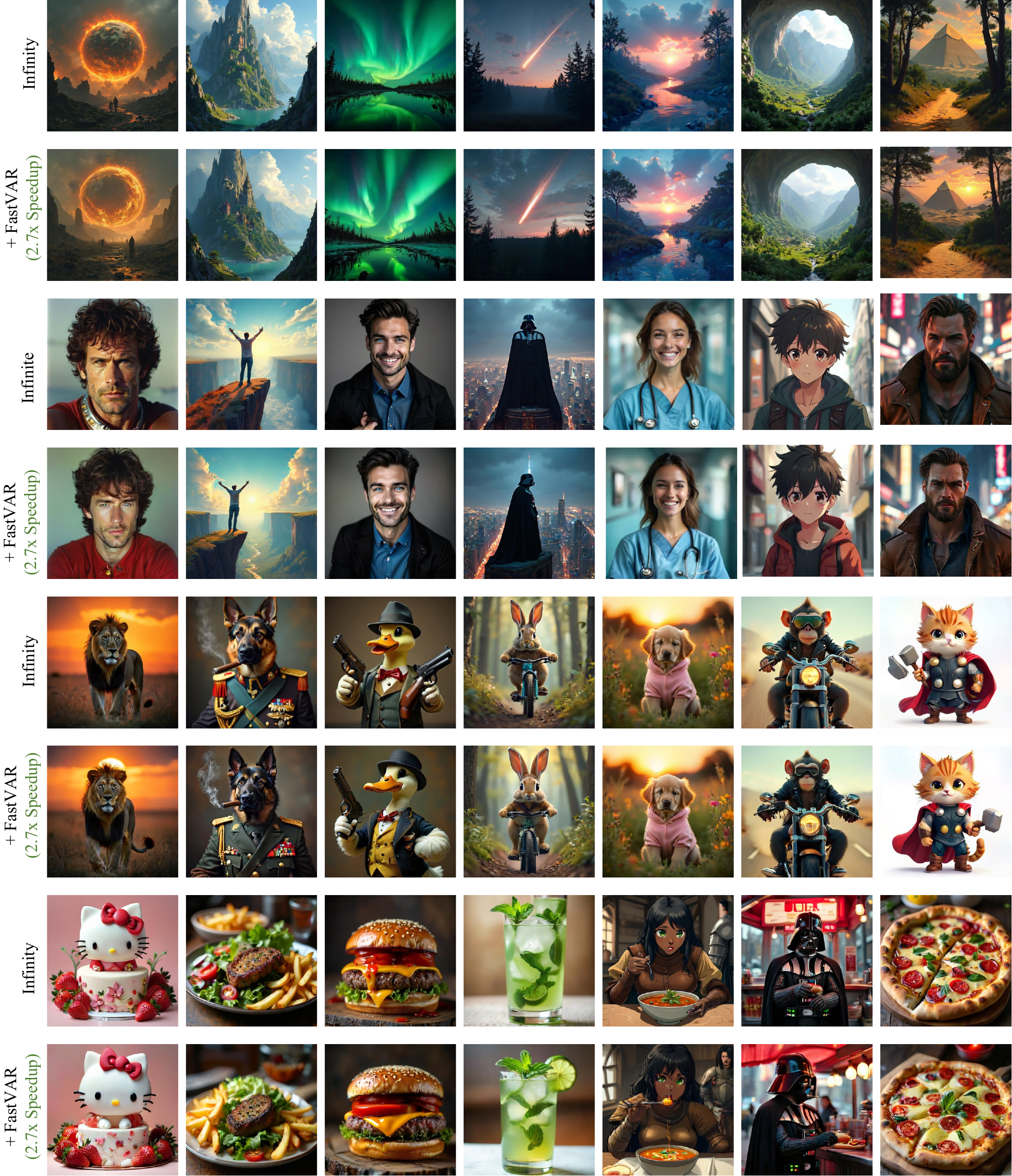}
    \vspace{-2mm}
    \caption{More qualitative comparison with the Infinite model on the MJHQ30K dataset.}
    \label{fig:suppl-mjhq30k-infinite-compare}
\end{figure*}

\begin{figure*}[ht]
    \centering
    \vspace{-2mm}
    \includegraphics[width=0.96\linewidth]{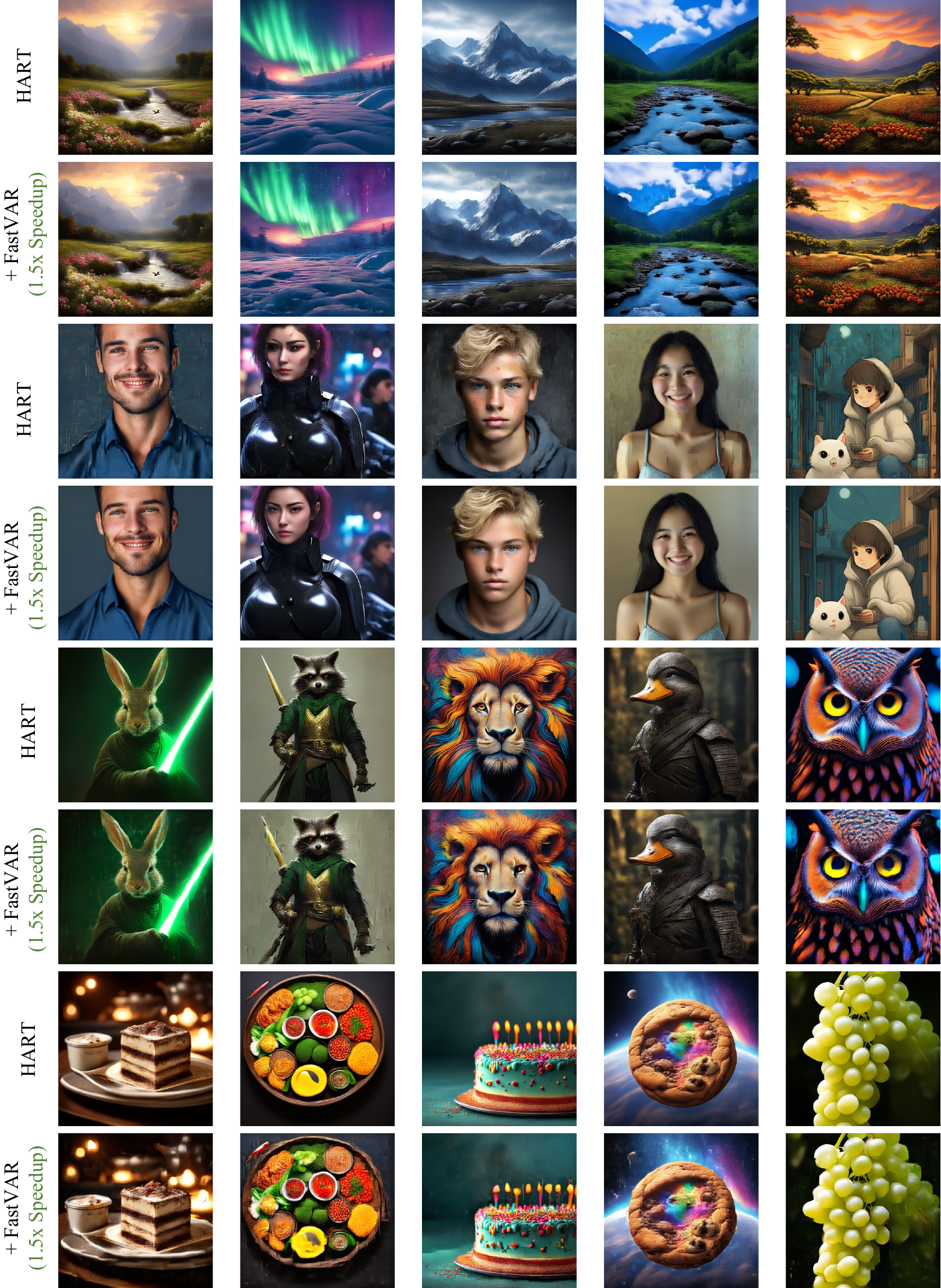}
    \vspace{-2mm}
    \caption{More qualitative comparison with the HART backbone on the MJHQ30K dataset.}
    \label{fig:suppl-mjhq30k-hart-compare}
\end{figure*}

\begin{figure*}[ht]
    \centering
    \includegraphics[width=0.96\linewidth]{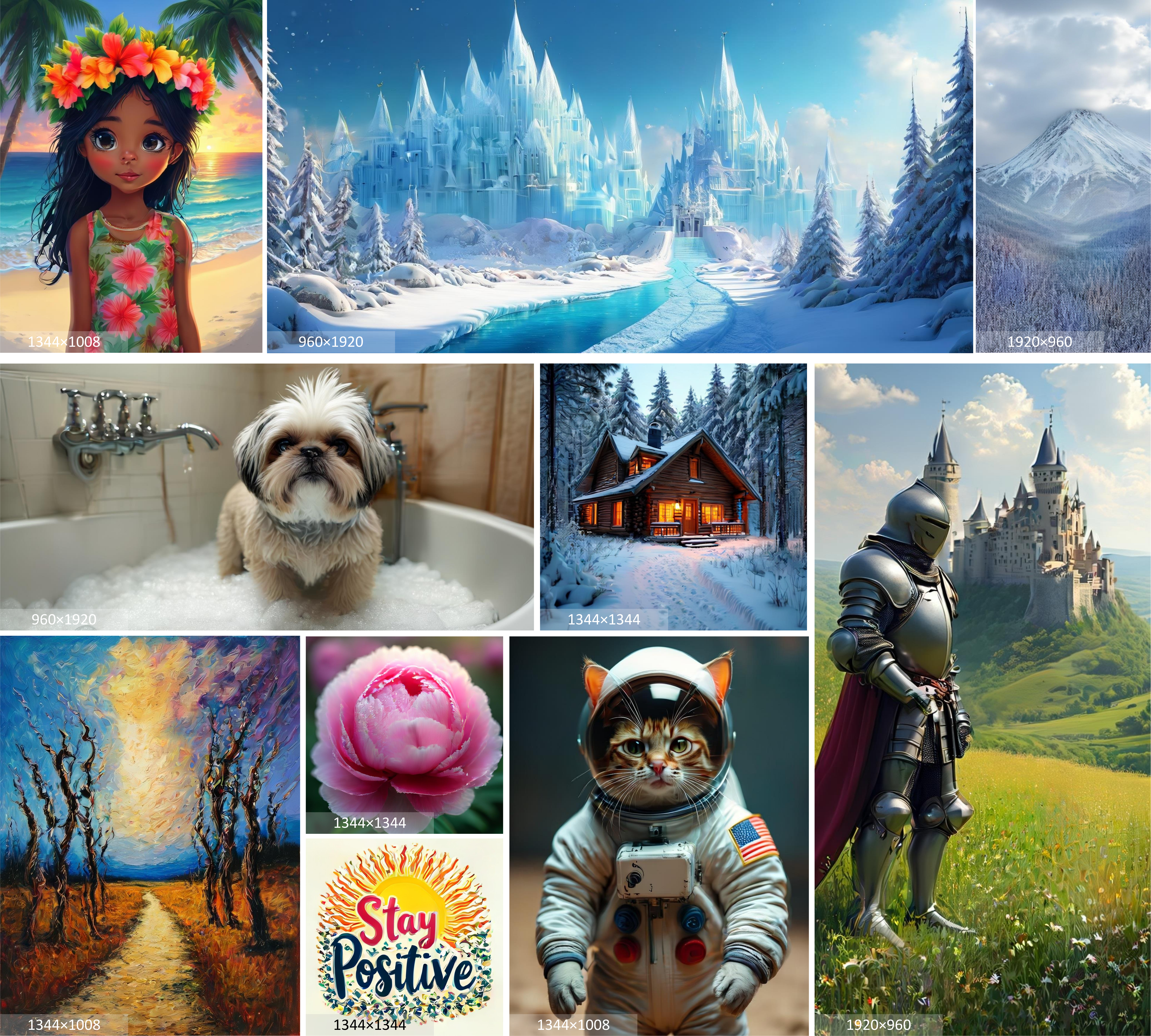}
    \vspace{-2mm}
    \caption{Moreover generation results of the high-resolution image synthesis with Infinite+FastVAR. The images are scaled for better presentation.}
    \label{fig:suppl-high-resolution-1}
\end{figure*}

\begin{figure*}[ht]
    \centering
    \includegraphics[width=0.67\linewidth]{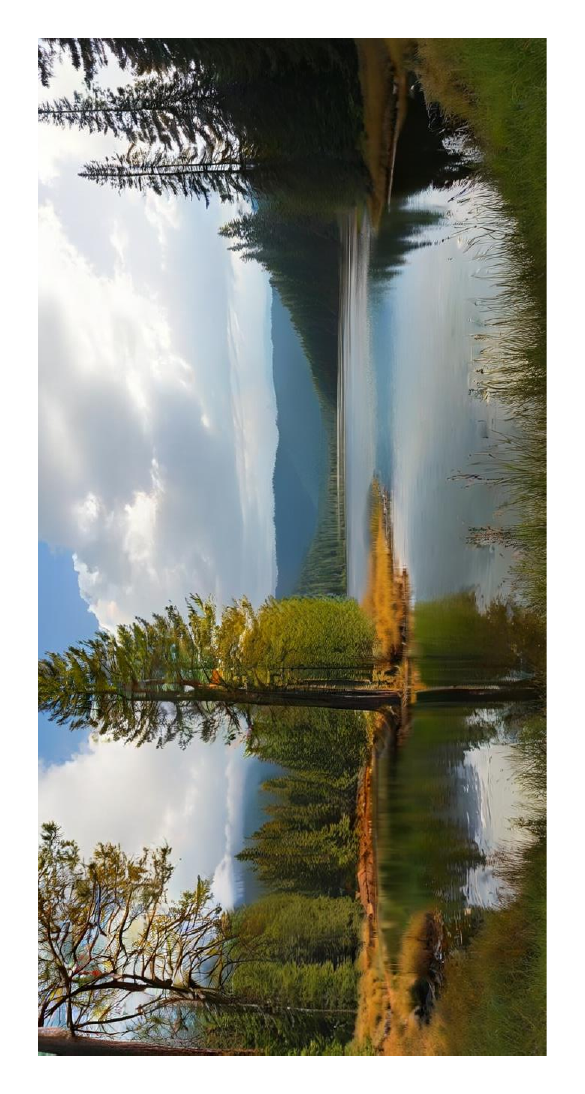}
    \vspace{-2mm}
    \caption{Moreover generation results of the high-resolution image synthesis with Infinite+FastVAR.}
    \label{fig:suppl-high-resolution-2}
\end{figure*}

\begin{figure*}[ht]
    \centering
    \includegraphics[width=0.67\linewidth]{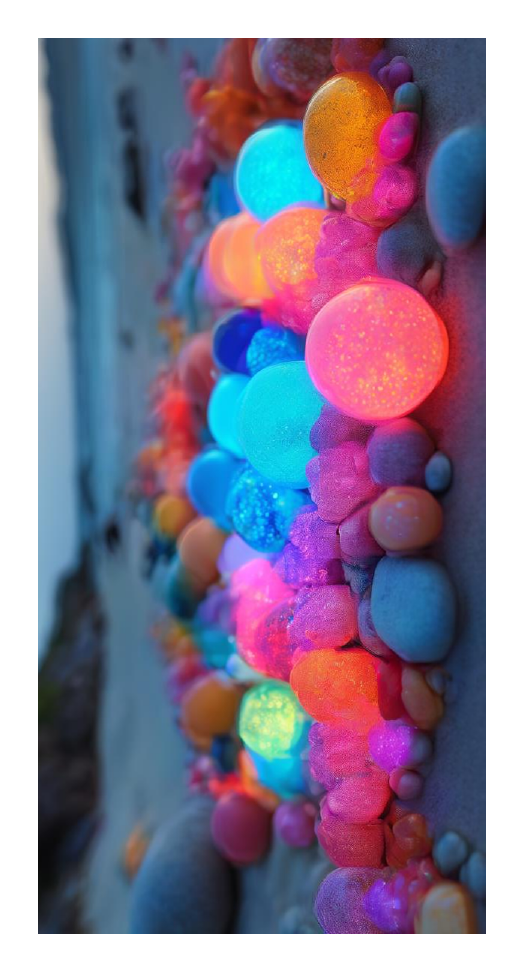}
    \vspace{-2mm}
    \caption{Moreover generation results of the high-resolution image synthesis with Infinite+FastVAR.}
    \label{fig:suppl-high-resolution-3}
\end{figure*}

In this section, we provide more visual results, which are organized as follows:

\noindent \begin{itemize}
     \item In~\cref{fig:suppl-intermidiate}, we give more visualization results about the intermediate outputs of the pre-trained VAR model at different scale steps.
     \item In~\cref{fig:suppl-mjhq30k-infinite-compare}, we give more qualitative results of Infinite~\cite{han2024infinity} on the MJHQ30K dataset.
    \item ~\cref{fig:suppl-mjhq30k-hart-compare} gives more qualitative results of the HART~\cite{tang2024hart} model on the MJHQ30K~\cite{li2024mjhq} dataset.
    \item In~\cref{fig:suppl-high-resolution-1}, ~\cref{fig:suppl-high-resolution-2}, and~\cref{fig:suppl-high-resolution-3}, we give more generation results on the zero-shot higher-resolution image synthesis tasks. 
\end{itemize}

\clearpage

\clearpage

\clearpage

\clearpage

\clearpage

\end{document}